\documentclass[10pt,twocolumn,letterpaper]{article}

\usepackage{cvpr}
\usepackage{times}
\usepackage{epsfig}
\usepackage{graphicx}
\usepackage{amsmath}
\usepackage{amssymb}
\usepackage{authblk}
\usepackage[dvipsnames]{xcolor}

\usepackage[noend]{algpseudocode}
\usepackage[ruled,vlined]{algorithm2e}
\usepackage{booktabs}
\usepackage{comment}
\usepackage[implicit=false]{hyperref}
\cvprfinalcopy 

\def\cvprPaperID{****} 

\begin{document}

\title{Self-Guided Adaptation: Progressive Representation Alignment for Domain Adaptive Object Detection}

\author{
 {Zongxian Li$^{1,4}$, Qixiang Ye$^{2,4}$, Chong Zhang$^{1,3,4}$, Jingjing Liu$^{5}$, Shijian Lu$^{4,6}$, Yonghong Tian$^{1,4}$}
 \\
 {$^{1}$ NELVT, School of EE \& CS, Peking University}
 {$^{2}$ University of Chinese Academy of Sciences} \\
 {$^{3}$ School of ECE, Peking University}
 {$^{4}$ Pengcheng Laboratory} \\
 {$^{5}$ Beijing Institute of Technology}
 {$^{6}$ Nanyang Technological University}
 
}
\maketitle

\begin{abstract}
		Unsupervised domain adaptation (UDA) has achieved unprecedented success in improving the cross-domain robustness of object detection models. 
		%
However, existing UDA methods largely ignore the instantaneous data distribution during model learning, which could deteriorate the feature representation given large domain shift.
In this work, we propose a Self-Guided Adaptation (SGA) model, target at aligning feature representation and transferring object detection models across domains while considering the instantaneous alignment difficulty.
The core of SGA is to calculate  ``hardness" factors for sample pairs indicating domain distance in a kernel space.
With the hardness factor, the proposed SGA adaptively indicates the importance of samples and assigns them different constrains. Indicated by hardness factors, Self-Guided Progressive Sampling (SPS) is implemented in an ``easy-to-hard'' way during model adaptation. 
Using multi-stage convolutional features, SGA is further aggregated to fully align hierarchical representations of detection models. 
Extensive experiments on commonly used benchmarks show that SGA improves the state-of-the-art methods with significant margins, while demonstrating the effectiveness on large domain shift.
\end{abstract}

\section{Introduction}
Convolutional neural networks (CNNs)~\cite{krizhevsky2012imagenet} have become one prevalent model for computer vision tasks, such as image classification~\cite{krizhevsky2012imagenet,simonyan2014very} and object detection~\cite{girshick2014rich,ren2015faster,liu2016ssd}. 
Nevertheless, many CNN models, $e.g.,$ CNN-based object detectors, require a large amount of annotated training data, which are costly and time-consuming to collect. 
Transferring detection models trained on a label-rich domain (publicly annotated datasets) to an unlabeled domain (real-world scenarios) in an unsupervised way has therefore attracted increasing interests recently~\cite{chen2018domain,saito2019strong,zhu2019adapting,wang2019towards}.

\begin{figure}
	\centerline{\includegraphics[width=1\columnwidth]{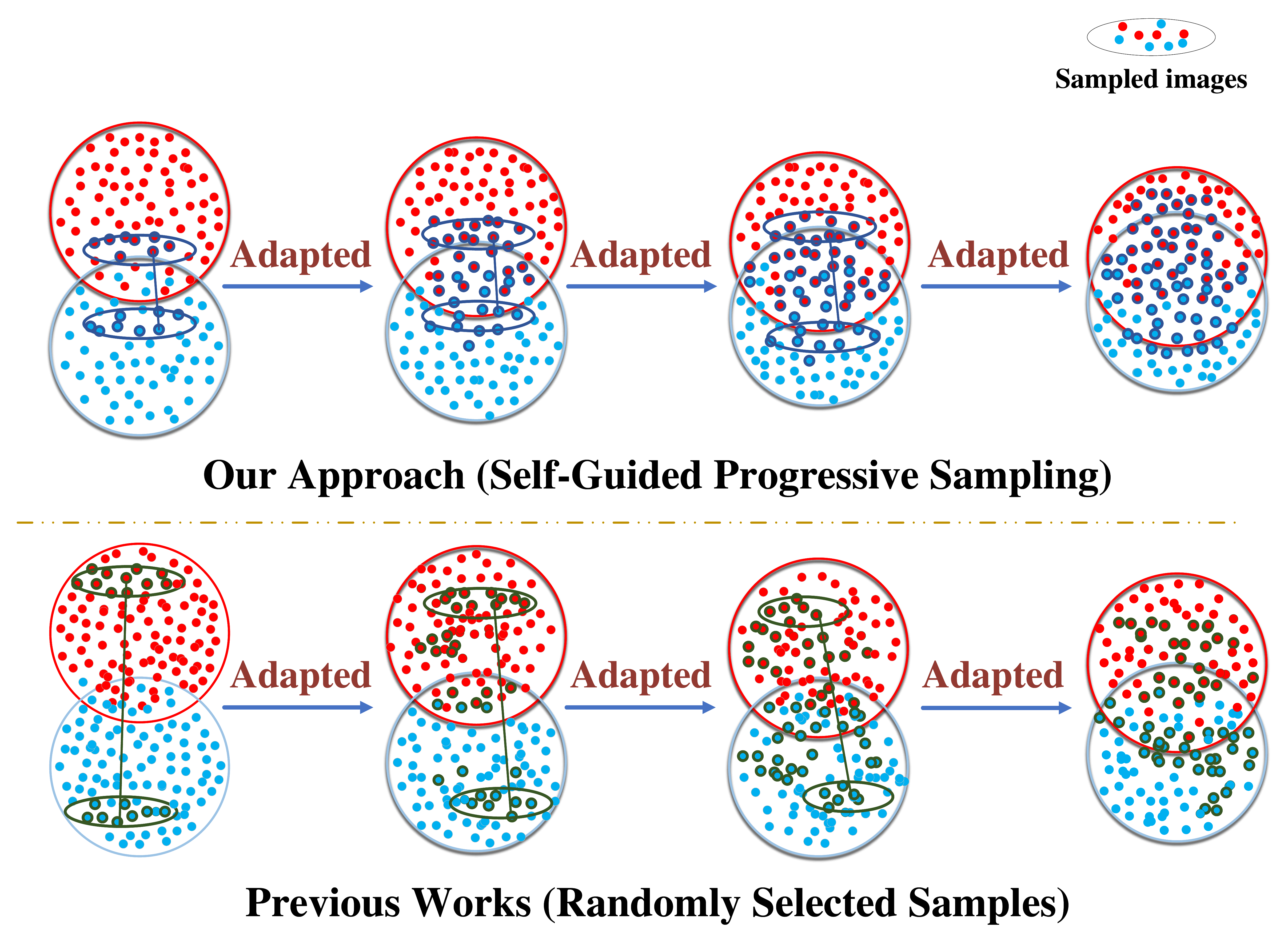}}
	\caption{Self-Guided Adaptation (SGA) approach adaptively samples image pairs around domain boundary, so that the adaptation procedure is implemented in a progressive "easy-to-hard" manner(top). In contrast, conventional approaches randomly select samples for while ignoring the sample distribution during adaptation (down).}
	\label{figure:motivation}
\end{figure}
As the core of transfer learning, unsupervised domain adaptation (UDA) has been extensively explored for model and feature transfer ~\cite{pan2009survey,pan2010domain,long2015learning,tzeng2015simultaneous}.
Early UDA methods mainly focused on the field of image recognition, while recently shifted to the object detection~\cite{chen2018domain,zhu2019adapting}. 
	The well-known Faster R-CNN~\cite{ren2015faster} has been updated to be domain adaptive by using adversarial learning to align feature between source and target domains.
	And then, ~\cite{saito2019strong} and ~\cite{zhu2019adapting} attempted to reduce domain discrepancy respectively at global- and region-level. 
	%
	
	In a broad view, most UDA methods achieved cross-domain feature alignment root on the adversarial learning while they largely ignore the sampling strategy for each mini-batch when optimizing domain adaptation models. 
	It is therefore unreasonable to align feature representations under a fixed constraint without considering the instantaneous domain shift.  Accordingly, immutable sampling strategies which assume that each sample is of equal importance for adaptation is implausible. Considering that the sample distribution is dynamic, $e.g.,$ some source domain samples have been falling into the target domain (easy-to-align) while others are far from the target domain (hard-to-align), Fig.\ \ref{figure:motivation}.  
	In addition, \cite{zhu2019adapting,chen2018domain,cai2019exploring} attempted to align feature distribution on the generated region candidates. Nevertheless, directly aligning at instance level is implausible as it is hard to generate precise region proposals if there exists large domain shift.
	Recent research \cite{saito2019strong} attempted to solve the problem by introducing the Focal Loss~\cite{lin2017focal} to weight and reduce the impact of hard samples. Nevertheless, the weights applied on training samples are still fixed, and can not be adaptive to the change of domain shift and variation of the sample distribution.

	In this study, we propose a Self-Guided Adaptation (SGA) model with a Self-Guided Progressive Sampling(SPS) strategy, and target at aligning feature representation and transferring detection models across domains while considering the instantaneous alignment difficulties. Our SGA with SPS is inspired by the self-paced curriculum, which simulates the learning process of humans and gradually proceeds from easy to complex samples, and progressively aligning representation~\cite{SPCL2015} with fully respect to the instantaneous domain shift.

	The progressive representation alignment is implemented with adversarial learning by fully considering the hardness of sample pairs (one from the source domain and the other from the target domain) to be aligned. The hardness for each sample pair is defined on the feature distance between its two samples in a Reproducing Kernel Hilbert Space (RKHS). According to the hardness, we dynamically adjust the constraint for adversarial learning and implement domain adaptation in a progressive ``easy-to-hard" manner. During the learning procedure, the model tends to align easy sample pairs at early iterations and gradually shifts to hard ones at later iterations, Fig.~\ref{figure:motivation} (upper). 
	
	The contributions of this work are summarized as follows: 
	\begin{itemize}
		\item A Self-Guided Adaptation model (SGA), which implements the representation alignment dynamically by fully leveraging the instantaneous sample hardness defined in the Reproducing Kernel Hilbert Space (RKHS). 
		
		\item A Self-Guided Progressive Sampling (SPS) strategy based on the instantaneous sample hardness, which is able to leverage instantaneous sample distances for progressive sampling and representation alignment.
		
		\item State-of-the-art performances on commonly used benchmarks and significant effectiveness over various domain shift settings.
	\end{itemize}
%
\begin{figure*}[t]
	\centering
	\includegraphics[width = 1\linewidth]{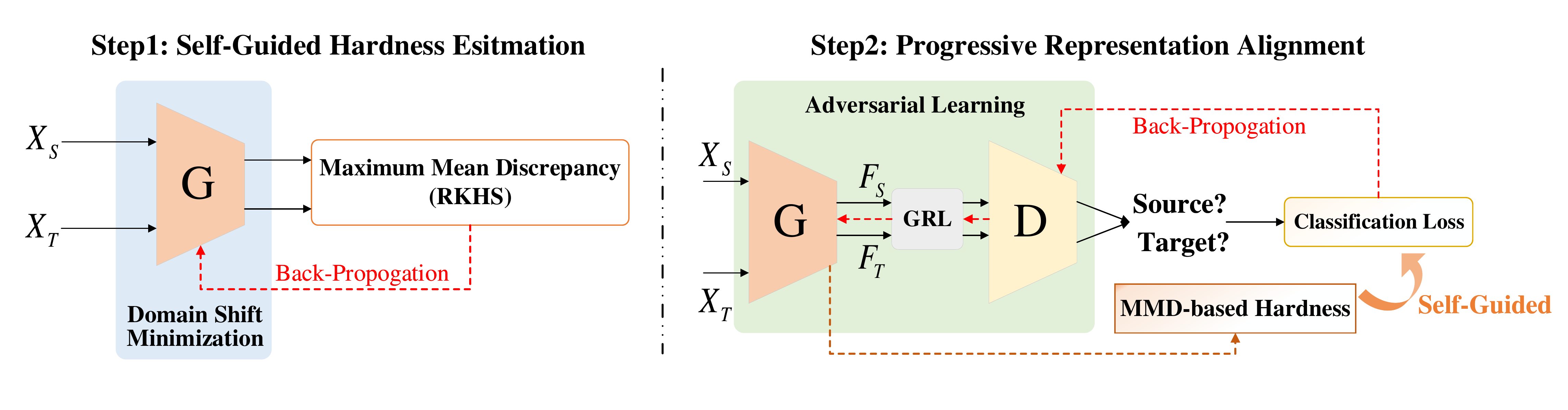}
	\caption{Self-Guided Adaptation (SGA) model. In the first step, the Maximum Mean Distance is calculated over each mini-batch in the RKHS, which indicates the hardness to be aligned of a pair of samples ($x_s$, $x_t$) from source and target domain. In the second step, a hardness-guided loss is designed for the domain discriminator $D$, with which we learn domain-adpative feature generator $G$ in a progressive and adversarial manner.}
	\label{figure:model}
\end{figure*}

\section{Related Work}

In this section, we first review UDA methods from a general perspective. We then review domain adaptive object detection approaches.
	
	\textbf{Unsupervised Domain Adaptation (UDA).}
	UDA aims to minimize the performance drop when transferring the model trained on the label-rich domain/dataset to an unlabelled domain/dataset.
	In the past few years, UDA has been extensively explored in broad fields of computer vision, including object classification~\cite{li2017domain, motiian2017unified, panareda2017open, long2015learning}, object detection~\cite{chen2018domain, saito2019strong, zhu2019adapting, kim2019diversify},and person re-identification~\cite{deng2018image,zhong2019invariance, pang2018cross}. 
	The key of UDA is to align the feature distributions of source and target domains. To this end, theoretical analysis about the domain/datasets shift are given in~\cite{pan2009survey,pan2010domain,long2015learning}, by measuring the feature distance between different domains/datasets~\cite{ganin2014unsupervised,long2015learning}.

	Based on the analysis, one line of UDA methods align feature representations by minimizing the domain distance.
	Maximum Mean Discrepancy (MMD)~\cite{gretton2012kernel}, a domain distance metric, was proposed to minimize the domain shift in the space of Reproducing Kernel Hilbert Space(RKHS)~\cite{long2015learning, long2016unsupervised, chopra2013dlid}. 
	The other line of methods \cite{DBLP:conf/icml/HoffmanTPZISED18, DBLP:conf/cvpr/BousmalisSDEK17,DBLP:conf/nips/LiuT16} attempted reducing the domain shift by taking advantages of adversarial discrimination to confuse source and target domains while aligning feature distributions. 
	The representative CyCADA \cite{DBLP:conf/icml/HoffmanTPZISED18} transferred samples across domains at both pixel- and feature-level.
	Domain confusion loss \cite{DBLP:journals/jmlr/GaninUAGLLML16,DBLP:journals/corr/AjakanGLLM14} was designed to learn domain-invariant features.
	Saito \emph{et al.} \cite{Saito_2018_CVPR} aligned distributions of source and target domains by maximizing the discrepancy of classifiers' outputs.
	In addition, training adaptation model with pseudo labels has achieved increasing attention.~\cite{chen2019progressive} achieved progressive alignment by assigning pseudo-labels to easy samples with respect to the intra-class distribution variance.
	
	\textbf{Domain Adaptive Object Detection.}
	UDA has attracted renewed interests in the object detection area since 2018~\cite{chen2018domain} with the key idea to align the feature distributions between source and target domains. Kuniaki~\textit{et al.} pointed out that more emphasis should be put on images that are globally similar and proposed the strong-weak distribution alignment~\cite{saito2019strong}. 

	Different with image classification that considers a holistic image, domain adaptive object detection focuses on local regions~\cite{zhu2019adapting}. Strongly matching the entire distributions of source
	and target images to each other at image level may fail, as domains have distinct scene layouts and different combinations of objects. 
	A Domain Adaptive Faster R-CNN(DA-Faster R-CNN)~\cite{chen2018domain} was proposed to minimize the discrepancy among two domains by exploring both image- and instance-level domain classifier in an adversarial manner. The similar motivation was used to align feature representation across domains on enlarged positive regions~\cite{zhu2019adapting}.
	Mean Teacher with object relations~\cite{cai2019exploring} was also considered, which addressed the adaptive detection from the viewpoint of graph-structured consistency.
	Moreover, reducing the domain gap by bridging an intermediate domain between source and target domain was adopted by~\cite{kim2019diversify,hsu2019progressive}.

	Many existing works extensively investigated the global- or region-level representation alignment, while unfortunately ignored the instantaneous alignment difficulty and sampling strategy during model learning, which could deteriorate the feature representation given a large domain shift. In this work, we propose a Self-Guided Adaptation strategy, and target at aligning representations and transferring models across domains while considering instantaneous sample distances. This different yet novel perspective makes our approach be complementary to many existing domain adaptive approaches.


\section{The Proposed Approach}
	In this section, we first describe the Self-Guided Adaptation (SGA) model based on sample hardness and adversarial learning, Fig.~\ref{figure:model}. We then detail the Self-Guided Progressive Sampling (SPS) for progressive representation alignment based on the proposed SGA. The entire procedure of representation alignment is organized by simulating the learning process of human and gradually proceeds from ``easy-to-hard'' samples. The two-stage object detector, Faster R-CNN, is employed as a base detector, and the alignment is operated on the backbone network, which includes features from three convolutional stages.

\subsection{Self-Guided Adaptation Model}

\begin{figure*}
	\centering
	\includegraphics[width = 1\linewidth]{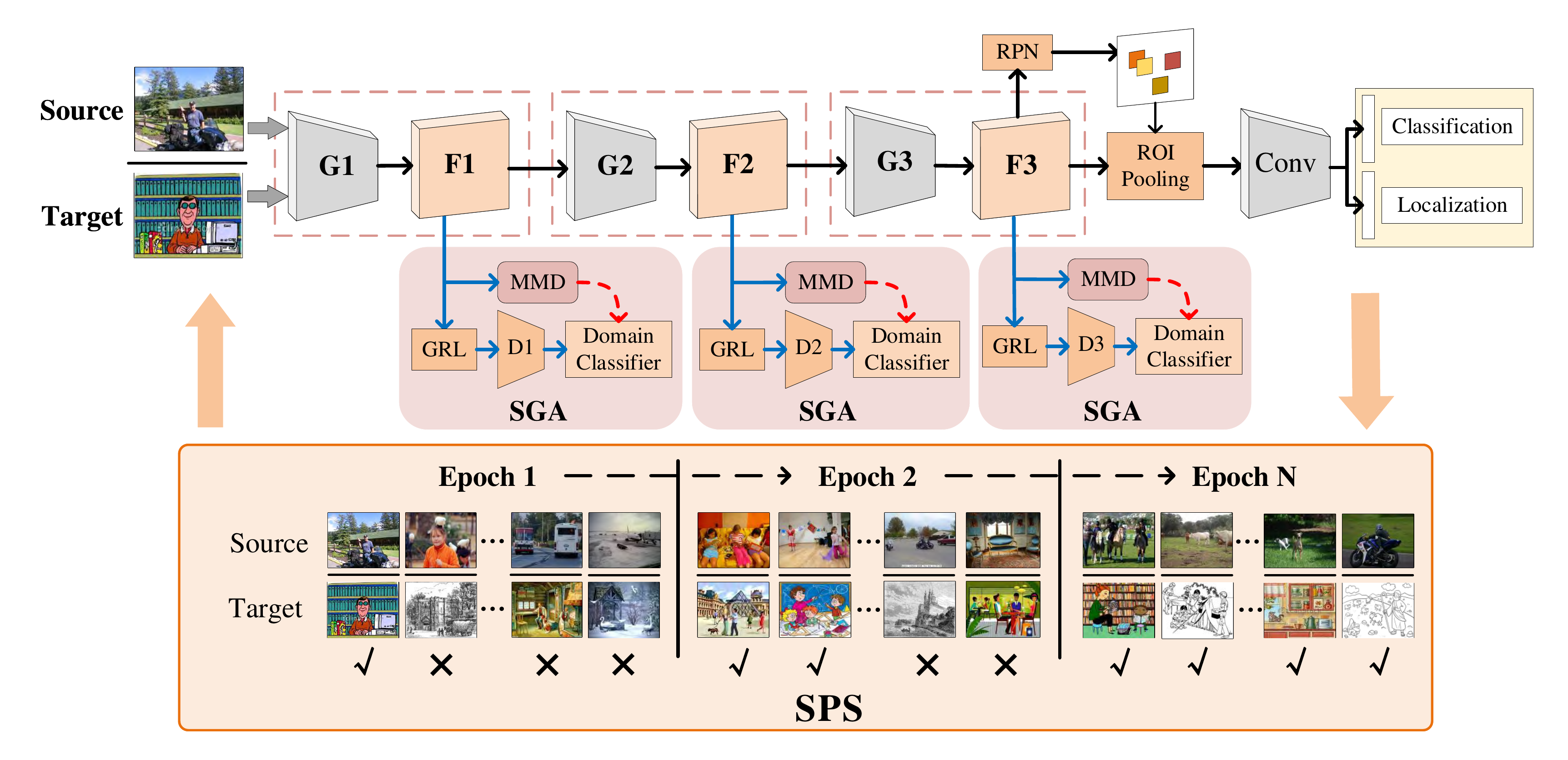}
	\caption{Illustration of the proposed approach for domain adaptive object detection. Three Self-Guided Adaptation (SGA) models are applied on three convolutional stages for representation alignment. The Self-Guided Progressive Sampling (SPS) adaptively selects samples so that the entire adaptation procedure is implemented in a progressive easy-to-hard manner. $\checkmark$ indicates selected samples and $\times$ discarded samples.}
	\label{figure:flowchart}
\end{figure*}
	In the UDA setting, a pair of samples is composed of a labeled image $(x^s, b^s, c^s)$ with fully supervision~(\textit{i.e.}, bounding boxes and categories) from source domain $X^S$, and an unlabeled image $(x^t)$ without any supervision from the target domain $X^T$. 
	The objective of SGA is performing model adaption from $X^S$ to $X^T$ while considering the instantaneous domain shift. To fulfill this purpose, a hardness factor is defined for each sample pair based on the Maximum Mean Discrepancy (MMD) distance in each mini-batch. 

\textbf{Self-Guided Hardness.} 
As a classical metric for comparing distributions based on the Reproducing Kernel Hilbert Space (RKHS)~\cite{gretton2012kernel}, denoted as $\mathcal{H}$, MMD has been widely used for minimizing the domain shift in the field of transfer learning, which is able to preserve all of the statistical features of arbitrary distributions by embedding distributions into infinite-dimensional feature space, while allowing one to compare and manipulate distributions using Hilbert space inner product operation~\cite{song2013kernel}.
	
	Considering two distributions $S$ and $T$, which respectively represent the source and the target domains, MMD is defined as
\begin{equation}
\setlength{\abovedisplayskip}{4.5pt}
\setlength{\belowdisplayskip}{4.5pt}
\begin{aligned}
{\gamma}(\mathfrak{F}, S, T) = sup_{f\in\mathfrak{F}}(\mathbb{E}[f(x^s)] - \mathbb{E}[f(x^t)]),
\end{aligned}
\end{equation}
	where $\mathfrak{F}$ denotes a set of functions of unit balls in a RKHS. As shown in Fig.\ \ref{figure:model} (left), two sample images are fed to a feature extractor ($G$) to extract features, based on which the instantaneous MMD is computed in each learning iteration.

	Denote $F_i^s$ and $F_i^t$ as the $i_{th}$ output features in a mini-batch and the space $\mathcal{H}$ is a Hilbert space with inner product $\left \langle\cdot,\cdot\right\rangle_{\mathcal{H}}$ and the corresponding norm $\small\Arrowvert\cdot\small\Arrowvert_{\mathcal{H}}$, the empirical estimate of MMD can be rewritten as
\begin{equation}
\setlength{\abovedisplayskip}{4.5pt}
\setlength{\belowdisplayskip}{4.5pt}
\begin{aligned}
	\gamma(F_i^s,F_i^t)=\bigg\Arrowvert\frac{1}{n_s}\sum_{i=1}^{n_s}\phi(F_i^s)-{\frac{1}{n_t}}\sum_{i=1}^{n_t}\phi(F_i^t)\bigg\Arrowvert_{\mathcal{H}},
\end{aligned}
\end{equation}
	where $\phi(\cdot)$ represents the kernel distance mapping: $\mathcal{X} \to \mathcal{H}$~\cite{gretton2012kernel}. $n_s$ and $n_t$ denote the numbers of source and target samples in a batch, respectively. According to~\cite{borgwardt2006integrating}, the $kernelized$ equation of the vector-matrix multiplication form of MMD is calculated as
\begin{equation}
\setlength{\abovedisplayskip}{4.5pt}
\setlength{\belowdisplayskip}{4.5pt}
\begin{aligned}
\gamma(F_i^s,F_i^t)=\bigg(&\frac{1}{n_s^2}\sum_{i=1}^{n_s}\sum_{j=1}^{n_s}k(F_i^s, F_j^s)\\ +  &\frac{1}{n_t^2}\sum_{i=1}^{n_t}\sum_{j=1}^{n_t}k(F_i^t, F_j^t) \\  -
&\frac{2}{n_{s}n_{t}}\sum_{i=1}^{n_s}\sum_{j=1}^{n_s}k(F_i^s, F_j^t)\bigg)^\frac{1}{2},
\label{k_MMD}
\end{aligned}
\end{equation}
where $k(\cdot) = RBF(\cdot)$ is a Radial Basis Function (RBF) kernel function.

	With Eq.\ \ref{k_MMD}, we calculate the MMD-based hardness for each sample pairs between source and target domains. Such hardness is used as a loss to minimize the domain shift as well as serving as a self-guided metric to determine the constraint of the representation alignment between sample pairs.
	Sample pair which owns a large MMD distance are hard to be aligned, and vice versa, and then we adaptively assign loss to sample pairs in an Self-Guided manner.

\textbf{Hardness-guided Adaptation.} 
	The adversarial learning framework is constructed by combining a feature generator ($G$) and a domain discriminator ($D$) with a Gradient Reverse Layer (GRL) module ~\cite{ganin2014unsupervised}. $G$ is used to extract domain-features while the $D$ requires to predict the probability of domain label $p$. $D$ is trained for distinguishing the source and target samples while $G$ is trained for deceiving the $D$ with the reversed gradient. In this way, $G$ tends to learn feature representation that can cover both the source and target domains.

	Based on the adversarial learning framework, we construct a hardness-guided model for representation alignment, as illustrated in Fig.~\ref{figure:model}(right) . 
	To introduce sample hardness into the adversarial procedure, we first update the domain classification loss of $D$ from the Cross-Entropy loss to the Focal loss~\cite{saito2019strong}, which assigns larger weights to easier samples (close to domain boundary) and smaller weights to harder ones (far away from domain boundary)\footnote{Sample pair will be considered as an easy one if it is hard to be classified under the domain adaptation setting, which is opposite of general classification.}.
%
Denote $p\in[0,1]$ as discriminator $D$'s estimated probability, the hardness-guided Focal Loss is defined as
\begin{equation}
\setlength{\abovedisplayskip}{5pt}
\setlength{\belowdisplayskip}{5pt}
\begin{aligned}
	L_{foc}(p_t,y) = (1-p_t)^{\gamma} log(p_t),  \\
\end{aligned}
\end{equation}
	The hardness factor $\gamma$ defined in RKHS is calculated by Eq.\ \ref{k_MMD}, constraining adversarial learning dynamically.
	$y$ denotes the domain label, which is assigned to 1 for a sample from source domain and 0 otherwise. 
	$p_t$ equals to the model's estimated probability $p$ with domain label $y=1$ and $p_t = (1-p)$ otherwise.
	Specifically for a pair of training samples $(x^s, x^t)$, the domain adaptive adversarial loss function is formulated as
\begin{equation}
\setlength{\abovedisplayskip}{5pt}
\setlength{\belowdisplayskip}{5pt}
\begin{aligned}
	L_{adv}(x^s, x^t) =& (1-D(G(x^s)))^{\gamma}log(D(G(x^s))) \\  
	& -(D(G(x^t)))^{\gamma}log(1-D(G(x^t))),  
\end{aligned}
\label{adv_loss}
\end{equation}
	which describes an adversarial learning procedure under the constraints of sample hardness $\gamma$. With the $\gamma$ estimated in the RKHS, the SGA adaptively indicates the importance of samples and assign them different constraints.
	
	Accordingly, the total loss function for a domain adaptive detector considering a batch of samples is concluded as 
\begin{equation}
\setlength{\abovedisplayskip}{5pt}
\setlength{\belowdisplayskip}{5pt}
\begin{aligned}
	L_{batch} = \frac{1}{n}\sum_{i=1}^{n}(L_{det}^{i} + L_{adv}^{i} + \beta L_\gamma^{i}),
\end{aligned}
\label{minibatch_loss}
\end{equation}
	where $n$ denotes the number of samples in a mini-batch. $L_{det}$ denotes the loss function of Faster R-CNN over training samples in the source domain. $L_{\gamma}$ denotes the loss about sample hardness, defined as $L_{\gamma}={\gamma}$. $\beta$ is a regularization factor which is experimentally determined, and we set $\beta=0.25$ for all experiments.

\subsection{Progressive Representation Alignment}

	The SGA model is defined for samples in each mini-batch. In what follows, we further propose a sampling strategy to construct mini-batches for progressive representation learning and alignment. 

\textbf{Self-Guided Progressive Sampling (SPS).}
	In conventional transfer learning approaches, sample pairs are randomly selected from source and target domains, without considering the difficulty of alignment and the sample distribution, as shown in the second row of Fig~\ref{figure:motivation}. 
	We argue that this sampling strategy is implausible as selected samples could have very large domain distances, which are difficult to be aligned at early iterations.

	Motivated by the self-paced learning paradigm~\cite{kumar2010self}, we propose to train the adaptation model with an ``easy-to-hard" way.
	Easy sample pairs that are ``easy-to-align'' will be selected with a higher priority in the early training iterations, while harder sample pairs will be selected later.
	To measure the alignment difficulty of training sample pairs, we use the average ``hardness"  calculated on samples among SGA modules once again.

	
	Specifically, we define a sampling strategy by introducing an adaptive threshold $\alpha$ over the "hardness". In each training iteration, sample pairs which have smaller  average ``hardness" than $\alpha$ are selected for model optimization. We first train the model for an pre-epoch and record the ``hardness" for each iteration and sort it,  and the ``hardness'' median value is selected as an initial $\alpha$. We retrain the model with the selected initial $\alpha$. After each training epoch, $\alpha$ is updated to a new median value according to the sorted recorded ``hardness" during the previous training epoch, which means that $\alpha$ is keep decreasing with model adapting, and more samples can be automatically included into training in a self-guided manner. Accordingly, the sampled loss function is defined as
\begin{eqnarray}
L = 
\begin{cases}
	v L_{batch},  \quad if \quad avg(\gamma) \leq \alpha \\
	0, \quad\quad\quad otherwise
\end{cases},
\label{total_loss}
\end{eqnarray}
\begin{algorithm} [t]
	\caption{Progressive Representation Alignment} 
	\KwIn{input pair: ($x^s$, $x^t$), initial threshold: $\alpha$} 
	$mmd\_ rec= [~]$\; 
	\For{$i=1;i \le n$, $n$ is the number of training epoch} 
	{ 
		\For{$j=1, j \le m$, $m$ is the number of steps}
		{
			$F_{i}^{s}=G(x_i^s), F_i^t=G(x_i^t)$\;
			Estimate Hardness $\gamma$ by Eq.\ \ref{k_MMD}\;
			Calculate adversarial loss $L_{adv}$ by Eq.\ \ref{adv_loss}\;
			Calculate mini-batch loss $L_{batch}$ by Eq.\ \ref{minibatch_loss}\;
			$mmd\_rec.append(avg(\gamma))$\;
			\If{$avg(\gamma) \leq \alpha$}
			{
				Calculate total loss $L$ by Eq.\ \ref{total_loss}\;
				$L.backward()$
			}
		}
		$mmd\_rec.sort()$\;
		$update~\alpha = median(mmd\_rec())$
	} 
	\label{alg.learning}
\end{algorithm}
	where $v$ determines whether or not a sample pair should be selected for aligning. $v=1$ if the average of the hardness factors satisfy $avg(\gamma) \leq \alpha$ and $v=0$, otherwise, where $avg(\cdot)$ refers to the average estimated hardness values among SGA modules. In this way, easy pairs contain instances with same categories or simillar appearance will be slected in the early iteration, as indicated in Fig.\ \ref{figure:flowchart}.
\begin{table*}[t]
	\small
	\setlength{\abovecaptionskip}{-10pt}
	\setlength{\tabcolsep}{1mm}
	\begin{center}
		\begin{tabular}{c|ccc|ccccccc}
			\toprule [1 pt]
			 Methods&\ H-G\ &\ H-L \ &\ SPS\ & bicycle & bird & car & cat & dog &  person & mAP (\%) \\
			\hline
			 Source-Only   &-&-&-& 69.4&47.1&39.2&33.5&21.4 & 58.1 & 44.78    \\
			
			  DA-Faster~\cite{chen2018domain}&-&-&-& 75.2 & 40.6 & 48.0 & 31.5 	& 20.6 & 60.0 & 45.98    \\
			
			  WST-BSR~\cite{kim2019self}&-&-&-& 75.6 & 45.8 & 49.3 & 34.1 & 30.3 & 64.1 & 49.87   \\
			
			  SW-Faster~\cite{saito2019strong}&-&-&-& 82.3 & 55.9 & 46.5 & 32.7 & 35.5 & 66.7 & 53.27\\
			
			   SW-Faster$^{\star}$\cite{saito2019strong}&-&-&-& 67.2 & 55.8 & 48.1 & 39.1 & 32.4 & 64.4 & 51.17   \\			
			\hline			
			  Baseline-A&-&-&-& 69.7 & 49.1 & 47.2 & 28.3 & 21.7& 60.5& 46.08    \\
			
			  Baseline-B&-&-&-& 66.4& 53.7 	& 43.8 	& 37.9 	& 31.9 	& 65.3 	& 49.83    \\
			
			  SGA-G&\checkmark&- & -& 79.5 & 48.4& 48.6& 38.1& \textbf{38.3 }&64.2 & 52.85  \\
			
			  SGA-L&\checkmark&\checkmark&-& 81.2 & 54.9	 & 48.7 & 37.9 	& 37.8 	& 66.0 	& 54.42 \\
			
			  SGA-S(Avg)&\checkmark&\checkmark &\checkmark& 87.7	 & 54.1 & 48.5 & 38.3 & 37.4& 65.2 &55.20\\
			  SGA-S(Best)&\checkmark&\checkmark &\checkmark & \textbf{88.5} & 54.5 & 49.1 & \textbf38.0 & 37.2& 64.9 &\textbf{55.30}\\
			\bottomrule
		\end{tabular}
	\end{center}
	\caption{Ablation studies and comparison on the Pascal VOC $\rightarrow$ WaterColor task: SW-Faster$^*$ donates our reproduction of the ~\cite{saito2019strong}. \textbf{SGA-S} is a complete implementation of our proposed method. \textbf{Baseline-A}, \textbf{Baseline-B}, \textbf{SGA-G}, and \textbf{SGA-L} are trained for ablation studies. H-G, H-L, and SPS denote the proposed hardness-guided adversarial loss $L_{adv}$, hardness loss $L_{\gamma}$ and Self-Guided Progressive Sampling, respectively.}
	\label{tab:pascal2watercolor}
\end{table*}


\textbf{Implementation.}
	Based on the SGA model and SPS strategy, we implement domain adaptive object detection based on the Faster R-CNN framework, Fig.~\ref{figure:flowchart}. 
	Given an image from the source domain, convolutional features are first extracted by the backbone network. Region proposal network (RPN) is used to generate proposals, and ROI pooling is used to extract features for object classification and localization. The model is trained in by optimizing object detection loss $L_{det}$ in the source domain. The objective of the proposed Self-Guided Adaptation is transferring the supervised detection model from source to target domain without using any annotation involved in the target domain. 
	We propose to apply three SGA modules corresponding three stages of features. Each module has an independent domain classifier at different stages of the backbone network, Fig. \ref{figure:flowchart}.
	Three SGA modules are simultaneously optimized with respect to each domain classifier.

%
	During each learning iteration, a mini-batch of samples are selected with the proposed SPS, which are used to adapt the learned detection model to the target domain by aligning feature representation of the source and the target domain in a Self-Guided manner. The entire learning procedure is summarized in Algorithm.\ \ref{alg.learning}. 

	Note that the proposed domain adaptation procedure is performed on feature maps instead of region proposals. The reason lies that the MMD-based hardness constraint is able to help finding image pairs with similar content and containing objects from the same categories to a certain extent. This alleviates the mismatch of representation adaptation across object categories and enables our approach to avoid relying on region proposals, which greatly simplifies the model adaptation procedure. Furthermore, directly aligning
	at instance level may fail since it is hard for RPN to generate precise region candidates if there exists large domain shift among feature maps which are not be well-aligned.

\section{Experiments}
	Experiments are conducted over four domain shift tasks including Pascal VOC~\cite{everingham2010pascal} $\rightarrow$ WaterColor\cite{inoue2018cross}, Cityscapes\cite{cordts2016cityscapes} $\rightarrow$ FoggyCityscape\cite{sakaridis2018semantic}, Cityscape $\rightarrow$ Detrac-Night\cite{lyu2018ua} and KITTI\cite{geiger2013vision} $\leftrightarrow$ Cityscape that have rich variations in domain shift caused by illumination conditions, cameras views, image styles, \textit{etc}. 
	We  compare the proposed approach with state-of-the-art methods and extensive ablation studies are conducted to validate the effectiveness of each proposed component.

\subsection{Experiments Settings}
	Faster R-CNN (ResNet101~\cite{he2016deep}-based) pre-trained on the ImageNet~\cite{deng2018image} is employed as the base detector in all experiments. While training the domain adaptation network, the inputs are a pair of images, including an annotated image from the source domain and an unannotated image from the target domain. The network is trained with a learning rate of 0.001 in the first 50,000 iterations and decreased to 0.0001 in the following iterations. All experiments are implemented by using the widely used Pytorch framework~\cite{paszke2017automatic}. 
	Without specific notations, we respectively report the average and the best mean Average Precision(mAP) observed from 70,000  to 100,000 iterations for evaluation and a fair comparison.

\begin{table*}[t]
	\small
	\setlength{\abovecaptionskip}{-10pt}
	\setlength{\tabcolsep}{1mm}
	\begin{center}
		\begin{tabular}{c|ccc|ccccccccc}
			\toprule [1 pt]
			Methods   &\ H-G\ &  \ H-L \  &\ SPS \ &  bus  &  cycle  &  car    &  bike   &   prsn   &   rider   &   train   &   truck   &    mAP (\%)     \\
			\hline
			\hline
			
			Source-Only   &-&-&- &  24.5  &  28.7  & 36.1  	& 19.9  &  25.7  &  33.4  &  10.5 & 19.8  & 24.83   \\
			
			\footnotesize{SW-Faster}~\cite{saito2019strong} &-&-&- & 36.2 & 35.3 & 43.5  &  30.0  &  29.9  &  42.3  &  32.6  & 24.5 &  34.29   \\   
			
			\footnotesize{SW-Faster$^*$}~\cite{saito2019strong} &-&-&- &  36.9  & 36.1 & 42.9  &  31.9  &  29.1  &  43.2 &  31.8  & 25.3 &  34.65   \\
			
			\footnotesize{Mean-Teacher}~\cite{cai2019exploring} &-&- &- &  30.6  	&  41.4  	&  44.0 		& 21.9 	& 38.6  & 40.6 	& 28.3 	& 35.6 	& 35.13  \\

			S-CDA~\cite{zhu2019adapting}  &-& -& -& 33.5 & 38.0 	& 48.5 & 26.5 	& 39.0 	& 23.3 	& 28.0 	& 33.6 	& 33.80  \\
			
			\hline
			Baseline-A   &-&- & -& 25.9 & 31.7 & 38.2 	& 22.6 	& 27.5 	& 24.1 	& 28.4 	& 26.1 	& 28.07  \\

			Baseline-B   &-&-& -& 34.7 	& 33.9 	& 42.4 	& 26.4 	& 27.8 	& 42.1 	& 15.7 	& 23.5 	& 30.81   \\
			
			SGA-G   &\checkmark&-& -& 43.7 & 32.5 & 44.1 & 25.6 & 29.6 & 43.8 & 32.1 & 23.4 & 34.35  \\
			
			SGA-L   &\checkmark&\checkmark &- & 46.6 & 33.1 & 43.8 & 22.7 & 30.2 & \textbf{44.3} & 37.3 & 26.1 & 35.51  \\
			
			SGA-S(Avg)  &\checkmark&\checkmark&\checkmark& 47.4 & 34.7 & 44.2 & 25.9 & 30.6 & 43.5 & 40.7 & 25.8 & 36.60    \\
			
 			SGA-S(Best)  &\checkmark&\checkmark&\checkmark & \textbf{51.6} & 35.1 & 44.5 & 26.4 & 31.9 & 43.2 & \textbf{41.3} & 29.5 & \textbf{37.94}  \\
			\bottomrule
		\end{tabular}
	\end{center}
	\caption{Comparison of SGA with state-of-the-art methods and ablation studies for the Cityscapes $\rightarrow$ FoggyCityscape task: SW-Faster$^*$ donates our reproduction of~\cite{saito2019strong} by using the ResNet101-based backbone.} 
	\label{tab:cityscapes2foggycityscapes}
\end{table*}

\begin{table}[t]
	\small
	\setlength{\abovecaptionskip}{-10pt}
	\setlength{\tabcolsep}{1mm}
	\begin{center}
		\begin{tabular}{c|ccc|c}
			\toprule [1 pt]
			\quad Methods &H-G& H-L & SPS &\quad AP on car(\%)\quad\\
			\hline
			\quad Source-Only\quad\quad &-&-&-&\quad42.02\quad\\
			\quad DA-Faster$^\star$~\cite{chen2018domain} &-&-&-&\quad44.51\quad \\
			\quad SW-Faster$^\star$~\cite{saito2019strong} &-&-&-&\quad46.43\quad \\
			\hline
			\quad Baseline-A\quad\quad &-&-&-&\quad44.12\quad\\
			\quad Baseline-B\quad\quad &-&-&-&\quad45.35\quad\\
			\quad SGA-G\quad\quad &\checkmark&-&-& \quad46.87\quad \\
			\quad SGA-L\quad\quad &\checkmark&\checkmark&-& \quad47.88\quad 	\\
			\quad SGA-S(Avg)\quad\quad &\checkmark&\checkmark&\checkmark&\quad48.59\quad\\
			\quad SGA-S(Best)\quad\quad &\checkmark&\checkmark&\checkmark&\quad\quad \textbf{49.67\quad} \\
			\bottomrule
		\end{tabular}
	\end{center}
	\caption{Comparison of our SGA with state-of-the-art methods as well as ablation studies for the daytime $\rightarrow$ night-time task: DA-Faster$^\star$ and SW-Faster$^\star$ denote our reproduction of the two methods by using the released code. } 
	\label{table:day2night}
\end{table}

\subsection{Domain Adaptive Object Detection}
	We compare the proposed SGA approach with a number of state-of-the-art works on adaption based object detection over four domain adaptation tasks,
	as listed in Tables.~\ref{tab:pascal2watercolor}-\ref{tab:kitti2cityscapes}. For each task, a model \textbf{Source-Only} is trained by using the annotated source domain images without any adaptation. 5 detection models are trained including: (1) \textbf{Baseline-A:} DA-Faster R-CNN~\cite{chen2018domain} with only image-level domain classifier trained with Cross-Entropy Loss; (2) \textbf{Baseline-B}: DA-Faster R-CNN with only image-level domain classifier trained with Focal Loss (the modulating factor $\gamma$ is fixed at 5); (3) \textbf{SGA-G}: the proposed SGA with hardness-guided adaptation modules and adversarial loss only ($i.e.,$ the first two terms in Eq.\ \ref{minibatch_loss}); (4) \textbf{SGA-L}: the SGA with both hardness-guided adversarial loss and hardness loss ($i.e.,$ all terms in Eq.\ \ref{minibatch_loss}); (5) \textbf{SGA-S}: the SGA incorporates all our designed hardness-guided adversarial adaptation loss,  hardness loss, and Self-Guided Progressive Sampling. The visualizations of object detection results over all domain adaptation settings are given in our supplementary document.
	
	\textbf{Natural Images to Artistic Images:} In this domain adaptation task, we use the training and validation splits of Pascal VOC 2007 and 2012~\cite{everingham2010pascal} as the source domain dataset and the WaterColor~\cite{inoue2018cross} as the target domain dataset. The source dataset Pascal VOC consists of around 15,000 images, while the WaterColor is collected from Behance website~\cite{wilber2017bam}, which consists of 1,000 artistic images that have the same 6 categories in common with the Pascal VOC. Images in two domains have very different styles, which bring a great challenge for domain adaptive object detection.

	As Table.~\ref{tab:pascal2watercolor} shows, the Source-Only which is trained by using source domain images without any adaptation does not perform well while applied to the target domain images. For this task, our proposed SGA-S obtains a superior mAP of 55.20\%, by sampling images from source and target domains progressively and train the model in a self-guided manner, which outperforms Baseline-A up to 9.12\%(from 46.08\% to 55.20\%) and  state-of-the-arts by large margins.  

	\textbf{Illumination Changes:} Illumination changes widely exists among images that are collected under different conditions in different environments. It is one of the most widely observed domain shifts, which often leads to a clear performance drop. We evaluate proposed approaches for two typical illumination change scenarios, namely, normal weather $\rightarrow$ foggy weather and daytime $\rightarrow$ nighttime.

	For the normal weather $\rightarrow$ foggy weather task, we adopt the  Cityscape~\cite{cordts2016cityscapes} as the source dataset and the FoggyCityscape ~\cite{sakaridis2018semantic} as the target dataset. Cityscape images capture almost all common traffic objects and FoggyCityscape is generated from Cityscape by adding fog noise. Both datasets consist of 2,975 training images and 500 validation images. Table.~\ref{tab:cityscapes2foggycityscapes} shows the comparison over the normal weather $\rightarrow$ foggy weather task. As Table.~\ref{tab:cityscapes2foggycityscapes} shows, the proposed SGA-S achieves superior performance over state-of-the-art methods(mAP: 36.60\% for average  \& 37.94\% for best), demonstrating the effectiveness of our method in handling dramatic weather condition changes. For a fair comparison, we update the backbone of SW-Faster~\cite{saito2019strong} from VGG16~\cite{Simonyan15} to ResNet101, donated as SW-Faster$^*$ in Table.~\ref{tab:cityscapes2foggycityscapes}.

\begin{table}[t]
	\small
	\setlength{\abovecaptionskip}{-10pt}
	\setlength{\tabcolsep}{1mm}
	\begin{center}
		\begin{tabular}{c|ccc|cc}
			\toprule [1 pt]
			Methods &H-G& H-L & SPS & K $\rightarrow$ C & C $\rightarrow$ K \quad \\
			\hline
			\quad Source-Only\quad &-&-&-&30.22&53.55\quad\\
			\quad DA-Faster~\cite{chen2018domain} &-&-&-&36.69&60.92\quad \\
			\quad S-CDA~\cite{zhu2019adapting} &-&-&-&42.70&- \quad\\
			\hline
			\quad Baseline-A &-&-&-&38.58&64.73\quad \\
			\quad Baseline-B &-&-&-&39.75   &68.17\quad \\
			\quad SGA-G &\checkmark&-&-&39.81	&68.92\quad \\
			\quad SGA-L &\checkmark&\checkmark&-&41.32	&69.55\quad\\
			\quad SGA-S(Avg)&\checkmark&\checkmark&\checkmark&{42.04}	&70.71\quad\\
			\quad SGA-S(Best) &\checkmark&\checkmark&\checkmark&\textbf{43.07}	&\textbf{71.43}\\
			\bottomrule
		\end{tabular}
	\end{center}
	\caption{Comparison of SGA with state-of-the-art methods as well as ablation studies over the KITTI $\rightleftharpoons$ Cityscapes task: Following the setting in~\cite{chen2018domain}, we report the AP on car in both adaptation direction, i.e. K $\rightarrow$ C and C $\rightarrow$ K.} 
	\label{tab:kitti2cityscapes}
\end{table}
	For the daytime $\rightarrow$ night-time task, the Cityscape~\cite{cordts2016cityscapes} is adopted as the source dataset and the Detrac-Night~\cite{lyu2018ua} is used as the target dataset. Detrac-Night is re-sampled from the UA-Detrac~\cite{lyu2018ua} dataset where images are captured in the night-time at different locations. From UD-Detrac, 3,500 images are selected for training and 500 images for testing. Table.~\ref{table:day2night} shows experimental results observed by training the model for 30,000 iterations. For state-of-the-art methods in ~\cite{chen2018domain,saito2019strong}, we run their released codes for fair comparisons.~\cite{saito2019strong} achieves very promising performance as it adopts a strong-and-weak alignment approach to capture domain-invariant features. As a comparison, the proposed SGA-S achieves superior performance (mAP: 48.59\% for average  \& 49.67\% for best).

	Domain shifts widely exist when images are collected by using different cameras with different resolutions and positioned at different viewpoints even when the image style and illumination conditions are similar~\cite{torralba2011unbiased}.
	Following the settings in~\cite{chen2018domain}, we conduct experiments for adaptation between the KITTI and the CityScape where images have very different resolutions and collected by different cameras. The results are observed by training the model for 30,000 iterations.

	As Table.~\ref{tab:kitti2cityscapes} shows, the proposed SGA-S significantly outperforms the DA-Faster R-CNN~\cite{chen2018domain} for adaptation in both directions. In addition, the region-based method~\cite{zhu2019adapting} achieves comparable performance as our approach for adaptation K $\rightarrow$ C. The close performance is largely attributed to the similar image styles between two domains where the misalignment is more about differences between the corresponding image regions.

\begin{table}[t]
	\small
	\setlength{\belowcaptionskip}{100pt}
	\setlength{\abovecaptionskip}{-10pt}
	\setlength{\tabcolsep}{1mm}
	\begin{center}
		\begin{tabular}{c|ccc|cc}
			\toprule [1 pt]
			Methods &Stage-1&Stage-2 & Stage-3 & mAP(Avg) & \quad mAP(Best) \quad \\
			\hline
			 SGA-S &-&-&\checkmark&35.12&36.58\quad \\
			 SGA-S &-&\checkmark&\checkmark&36.14   &37.22\quad \\
			 SGA-S &\checkmark&\checkmark&\checkmark&36.60	&37.94\quad \\
			\bottomrule
		\end{tabular}
	\end{center}
	\caption{Evaluation of SGA module numbers over the Cityscape $\rightarrow$ FoggyCityscapes.}
	\label{tab:SGA}
\end{table}

\subsection{Ablation Study}
	We perform ablation studies over the four domain adaptation tasks as shown in Tables.~\ref{tab:pascal2watercolor}-\ref{tab:kitti2cityscapes}. Additionally, we investigate the seperate contribution comes from each SGA module, as listed in Table.~\ref{tab:SGA}.

	Tables.~\ref{tab:pascal2watercolor}-\ref{tab:kitti2cityscapes} show that the proposed hardness-guided adversarial loss, hardness loss, and Self-Guided Progressive Sampling consistently improve the performance.
	%
	By training the model in an adversarial way with respect to the estimated hardness factor, SGA-G outperforms baselines by large margins in all four adaptation tasks.
	Take the Pascal VOC $\rightarrow$ WaterColor an example. The Baseline-A trained with classic cross-entropy loss obtains a marginal mAP improvement (+1.3\%) as compared with the Faster R-CNN (Source Only) which is trained by using the source domain images only. This indicates that direct alignment of representation without considering instantaneous domain shift is implausible.
	By capturing instantaneous domain shifts and assigning different losses to respective sample pairs, our SGA-G outperforms the Baseline-A and the Baseline-B by 6.77\% and 3.02\%(from 46.08\% and 49.83\% to 52.85\%), respectively. 

	The hardness loss which aims to minimize the domain shift clearly improves the detection performance across all four domain adaptation tasks. The ablation model SGA-L with the hardness loss outperforms the SGA-G from 0.63\% (for Cityscape $\rightarrow$ KITTI) to 1.57\% (for Pascal VOC $\rightarrow$ WaterColor) in mAP. The clear performance improvements are largely attributed to the hardness factor estimated in a RKHS which helps to preserve statistical features and close the domain gaps by jointly learning the domain-invariant features between the source and target domains and minimizing the domain shift, simultaneously. 
	The complete system SGA-S achieves the highest detection performances across all four domain adaptation tasks when we further sample image pairs from source and target domains progressively and train the model in a self-guided manner. The performance improvement over the SGA-L ranges from 0.71\% (for daytime $\rightarrow$ nighttime) to 1.22\% (for CityScape $\rightarrow$ KITTI). This clearly demonstrates the advantage of aligning samples from easy to hard progressively in a purely self-guided manner while training adaptive detection networks. 

	Table.~\ref{tab:SGA} demonstrates the seperate contribution from each SGA module over the domain shift: Cityscape$\rightarrow
	$FoggyCityscape. We observe that perfermances increase gradually when the SGA module is incorporated one by one, which backs up the effectivenss of aggregating hierarchical representation.

\subsection{Model Analysis}

\begin{figure*}[t]
	\centerline{\includegraphics[width=1\linewidth]{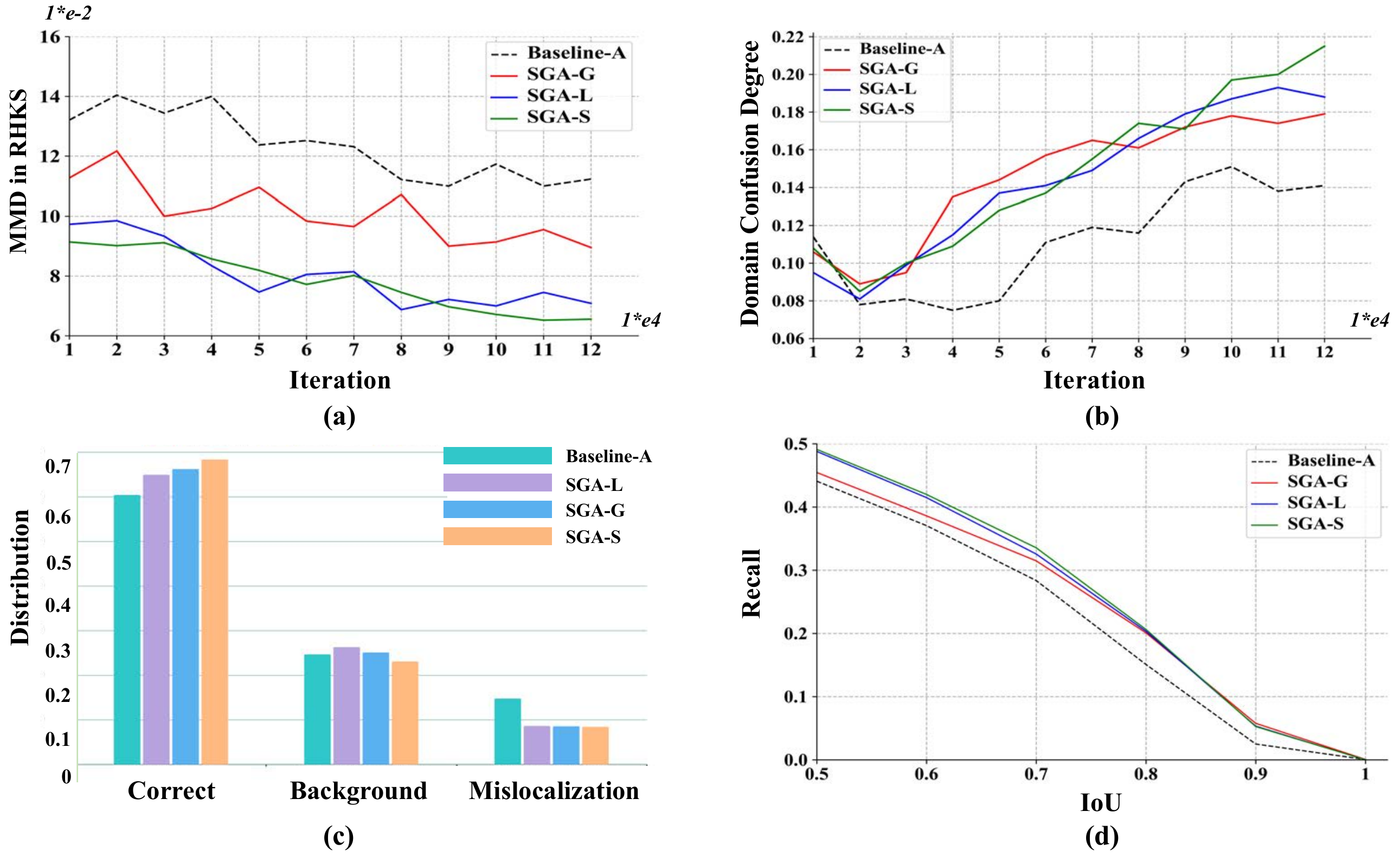}}   
	\caption{Model Analysis: (a) MMD-based hardness values for Baseline-A and our approaches; (b) Domain Confusion Degree analysis for Baseline-A and our approaches; (c) Error Analysis with respect to correct detection, background errors, and mislocalization, which based on 1,500 most confident detections. (d) Recall versus IoU threshold based on top-300 regions generated by RPN.}
	\label{figure:ErrorAnalysis}
\end{figure*}

	For further exploring the effectiveness of proposed approaches, we analyze both training procedures and detection results from following aspects, which are performed over tasks: Pascal VOC$\rightarrow$WaterColor${^{\ast}}$ and Cityscape$\rightarrow$FoggyCityscape$^{\star}$.

	\textbf{Qualitative Analysis on Hardness in RKHS$^{\ast}$ :} We compute the mean values of estimated hardness  by using Baseline-A and proposed methods, and illustrate the hardness values in Fig.~\ref{figure:ErrorAnalysis} (a).
	It can be seen that estimated MMD-based hardness among the source and target domain keeps decreasing by applying the proposed approach. The domain shift declines faster by optimizing the model with the "hardness" loss term. Moreover, SGA-S can further stabilize the decreasing procedure by selecting sample pairs in an ``easy-to-hard" manner.

	\textbf{Domain Confusion Degree$^{\ast}$ :} The degree is defined for better analyzing the domain confusion capacity of the proposed method. Specifically, an image from Source(Target) domain will be considered as a confused one, when it is wrongly classified as the Target(Source) image by $D$ with the classification probability $p$ satisfy $max(p) \leq 0.6$. 
	In Fig.~\ref{figure:ErrorAnalysis}(b), it can be clearly observed that our approach can better confuse samples from source and target domains.

	\textbf{Error Analysis on Top-ranked Detections$^{\ast}$:} We further diagnose proposed approaches by analyzing the detection errors among 1,500 top-ranked detections. Following the protocol in~\cite{chen2018domain}, the detection errors are categorized into three types including correct, mislocalization, and background error \footnote{Please refers~\cite{chen2018domain} for a clear definition about the three error types.}. As Fig.~\ref{figure:ErrorAnalysis} (c) shows, the correction detections increase gradually when our proposed approaches are incorporated one by one. At the same time, the mislocalization significantly drops when the self-guided components are incorporated. 

	\textbf{Recall Rate analysis$^{\star}$:} We evaluate recall vs. overlap for 300 top-ranked proposals generated by RPN, as shown in Fig.~\ref{figure:ErrorAnalysis}(d). The plot shows that proposed methods contribute a lot for generating high-quality region candidates at RPN stage. Specifically, with IoU of 0.7, SGA-S achieves 33.54\% recall, outperforming Baseline-A by 5.24 points, which strongly validates the effectiveness of hierarchical feature alignment at global feature level.

\section{Conclusion}

	In this paper, we propose a Self-Guided Adaptation (SGA) method, and target at aligning feature representation and transferring object detection models across domains in an adversarial way while considering the instantaneous domain shift. 
	To measure the domain shift, we design a ``hardness" factor for each sample pair in each mini-batch, indicating a domain distance in a kernel space.
	The hardness factor is further used as a metric to select training samples and achieve progressive representation alignment.
	With the proposed SGA and SPS, we implement the robust and effective domain adaptive object detection, improving the state-of-the-art methods with significant margins.
	The research in this paper not only demonstrates the effectiveness of the SGA upon domain adaptive object detection, but also provides a fresh insight to general UDA problems.

{\small
\bibliographystyle{ieee_fullname}
\bibliography{egbib}
}

\end{document}